# MedNet: Pre-trained Convolutional Neural Network Model for the Medical Imaging Tasks

**Laith Alzubaidi[1], J. Santamaría [2], Mohamed Manoufali[3], Beadaa Mohammed[4], Mohammed A. Fadhel[5], Jinglan Zhang[6], Ali H.Al-Timemy[7], Omran Al-Shamma[8], and Ye Duan[9]**

[1,6]School of Computer Science, Queensland University of Technology, Brisbane, QLD 4000, Australia

[2]Department of Computer Science, University of Jaen,´ 23071 Jaen,´ Spain

[1,3,4]School of Information Technology and Electrical Engineering, The University of Queensland, 4072, Brisbane, Australia

[5]College of Computer Science and Information Technology, University of Sumer, Thi Qar 64005, Iraq

[7]Biomedical Engineering Department, Al-Khwarizmi College of Engineering, University of Baghdad, Baghdad 10066, Iraq

[8]The University of Information Technology & Communications, Baghdad 10001, Iraq

[9]Faculty of Electrical Engineering& Computer Science, University of Missouri, Columbia, MO 65211, USA

## ABSTRACT

In the last few years, deep learning (DL) methods have dramatically improved the state-of-the-art in visual object recognition, speech recognition, and object detection, among many other applications. Specifically, DL requires a large amount of training data to provide quality outcomes. However, the field of medical imaging suffers from the lack of sufficient data for properly training DL models because medical images require manual labelling carried out by clinical experts thus the process is time-consuming, expensive, and error prone. Recently, transfer learning (TL) was introduced to reduce the need for the annotation procedure by means of transferring the knowledge performed by a previous task and then fine-tuning the result using a relatively small dataset. Nowadays, multiple classification methods from medical imaging make use of TL from general-purpose pre-trained models, e.g., ImageNet, which has been proven to be ineffective due to the mismatch between the features learned from natural images (ImageNet) and those more specific from medical images especially medical gray images such as X-rays. ImageNet does not have grayscale images such as MRI, CT, and X-ray. In this paper, we propose a novel DL model to be used for addressing classification tasks of medical imaging, called MedNet. To do so, we aim to issue two versions of MedNet. The first one is Gray-MedNet which will be trained on 3M publicly available gray-scale medical images including MRI, CT, X-ray, ultrasound, and PET. The second version is Color-MedNet which will be trained on 3M publicly available color medical images including histopathology, taken images, and many other. To validate the effectiveness MedNet, both versions will be fine-tuned to train on the target tasks of a more reduced set of medical images. MedNet performs as the pre-trained model to tackle any real-world application from medical imaging and achieve the level of generalization needed for dealing with medical imaging tasks, e.g. classification. MedNet would serve the research community as a baseline for future research.

## 1 INTRODUCTION

In the last two decades, machine-learning (ML) systems are increasingly being used to identify objects in images, transcribe speech into text, and match news items. Recently, these applications made use of a class of techniques called deep learning (DL) [7]. Specifically, DL allows computational models that are composed of multiple processing layers to learn representations of data with multiple levels of abstraction. DL methods have been extensively used and achieved improved results compared to previous research in many fields such as visual object recognition, speech recognition, and object detection, in addition to many other applications [7]. It is known that the performance of DL depends on a huge amount of data for representation learning [7]. The greater the amount of data supplied, the better the performance offered by DL. Usually, the available data are sufficient to obtain a good performance model, e.g. by using ImageNet dataset. However, the field of medical image analysis struggles from a lack of sufficient data for training DL models [8]. Moreover, medical images, such as that of ophthalmology or pathology slides, require manual subjective labeling and annotation from experts, which is time-consuming and prone to errors.

Training a convolutional neural networks (CNN) model from scratch will not provide significant outcomes, which is an additional drawback together with the lack of training data. In the last few years, transfer learning (TL) was introduced with the aim of dealing with the previous limitations [7]. Moreover, TL helps to obtain accuracy dealing with image classification tasks. TL is usually applied using a CNN model from ImageNet and its pre-trained weights. The model is then fine-tuned to tackle the target task efficiently. It was recently proven that the performance of the pre-trained models declined when they were employed for images such as chest x-rays and brain MRI, as these models have not been trained on medical gray-scale images [8]. Not only that, but a recent study has also shown that the performance of these models reduces when there is a small number of the target dataset. As a result, TL from these pre-trained models has not improved the performance of medical imaging applications [3].

The idea of training proposed MedNet on millions of medical images will improve the learned features better than that of the existing pre-trained model trained on usual natural images (ImageNet), so this would improve the performance when transferred to limited-data clinical problems, such as that of diseases diagnosis and predication.

In this paper, we aim at proposing a new model for addressing the previous drawbacks when dealing with classification tasks of medical images. The proposed model is named MedNet which will be in two versions namely Gray-MedNet and Color-MedNet. Both versions will be trained by using a publicly available dataset of 3M medical images for each version coming from multiple sources, e.g. MRI, CT, X-Ray, and PET, among many others.

## 2 STATEMENT OF THE PROBLEM

### 2.1 Transfer Learning

The lack of training data is a common problem for deep CNNs. The common solution in the state-of-the-art (SoTA) is using TL. TL employs the knowledge gained for a specific task to resolve other correlated tasks as well as to overcome the separated learning paradigm [7]. Also, the TL approach is based on the utilization of a trained network on different tasks for different source data then adjust it for the target task.

Hence, TL helps to obtain an improved rate of accuracy in image classification tasks by offering a large amount of labeled data, which is an extremely difficult task due to the effort and time taken to collect and label data. Thus, the latter issue guided us to motivate the use of TL and its proven outstanding performance.

### 2.2 Deep Learning and Transfer Learning Interaction

As stated, it is difficult to obtain accurate results by using DL due to the massive number of images required for training. Specifically, a deep CNN (DCNN) with many layers can achieve excellent results when dealing with image recognition and classification tasks, even better performance than a human in case an enormous volume of data is available [7]. However, these applications demand large datasets to prevent over-fitting and generalize DCNN models. Usually, there is no minimum size for the dataset in training a DCNN, but using small datasets or a DCNN with fewer layers would prevent the model to be highly accurate due to under or over-fitting issues. Then, models with fewer layers are less accurate because they are unable to use the hierarchical features of large datasets. On the other hand, collecting labeled datasets is extremely costly in fields such as environmental science and medical imaging [7]. In particular, in the field of medical image analysis, most crowd-sourcing workers do not have the required dataset for learning features [2,6].

Recently, many researchers have demonstrated that the use of TL in medical image classification tasks is effective and efficient [4,5,6]. Solving complicated issues in deep learning models requires a massive amount of data. Supervised models demand large medical/biological background to accurately annotate medical/biological images. Then, ML researchers frequently depend on the experts for annotation. Indeed, this is an unproductive and costly process. Thus, producing a significant amount of annotations to properly guide deep networks becomes impracticable.

**2.3 Overview of the State-of-The-Art**

In the past few years, researchers have been using several techniques to overcome the lack of training data. One of the most common techniques is data augmentation in which data is virtually created. Although such techniques enhance the data by creating further images, CNN models still struggle with overfitting issues due to repeated images in data augmentation. Recently, many researchers have employed a TL technique where DL models are trained using a large dataset, next fine-tuned to train on a smaller targeted dataset [6]. Nevertheless, although using TL improves the overall performance of CNN in many computer vision and pattern recognition tasks, it still has a fundamental challenge, which is the type of source data used for TL compared to the target dataset. Grosso modo, DCNN models trained using ImageNet, i.e. mainly comprised by natural images, are utilized to enhance the performance of complex medical image classification tasks. Additionally, the ImageNet dataset is quite different from the ones commonly used in medical imaging, which has a negative impact on the performance of the final result. It has been proven that different domain TL does not significantly affect the performance dealing with medical imaging tasks, e.g. lightweight models trained from scratch perform nearly as well as using the standard transferred models based on ImageNet [6,9]. There is a clear lack of a pre-trained model on medical images to help in learning, representation, and generalization. Therefore, this issue will be solved by the two versions of MedNet, introduced in the next section.

# 3 ARCHITECTURE OF MEDNET

MedNet is based on an effective DCNN model that combines several creative components to solve many issues, including feature extraction and representation, gradient-vanishing problem, and over-fitting. These components can be summarized as follows:

- Traditional convolutional layers at the beginning of the model to reduce the size of input images.

- Parallel convolutional layers with different filter sizes to extract different levels of features to guarantee that the model learns the small and large features.

- Residual connections and deep connections for better feature representation. These connections also handle the issue of gradient vanishing.

- Batch Normalization (BN) to expedite the training process.

- A rectified linear unit (ReLU) does not squeeze the input value, which helps minimize the effect of the vanishing gradient problem.

- Dropout to avoid the issue of over-fitting.

- Global average pooling makes an extreme dimensionality reduction by transforming the entire size to one dimension. This layer helps to reduce the effect of over-fitting.

The model, MedNet, starts with two traditional convolutional layers working in sequence. The first one has a filter size of 3×3, while the second convolution has a filter size of 5×5. Both convolutional layers are followed by BN and ReLU layers. Notice that we avoided utilizing small filters such as 1×1 at the beginning of the model. Then, it will prevent from losing small features, which will perform as a bottleneck. Eight blocks of parallel convolutional layers come after the traditional convolutional layers. Each block comprises four parallel convolutional layers with four distinct filter sizes (1×1, 3×3, 5×5, and 7×7). The output of these four layers is integrated into the concatenation layer to move to the following block. All convolutional layers in all eight blocks are followed by BN and ReLU layers. There are twelve connections between the blocks, some of them are short and others are long, with a single convolutional layer. These connections maintain the ability of the model to maintain different levels of features for the purpose of achieving a better representation of them. Both parallel convolutions and the connections are extremely important for gradient propagation as the error can back-propagate from multiple paths. Finally, two fully connected layers are adopted with one dropout layer between them. Softmax is employed to finalize the output. In total, MedNet consists of 44 convolutional layers.

## 4 MEDNET: TRAINING

We propose a novel technique of TL to overcome the issues of TL from pre-trained models of the ImageNet dataset when dealing with medical imaging tasks. Moreover,the proposed TL technique helps to address the issue of the lack of training data present in medical imaging applications and reduce the time of the annotation process. The proposed technique is based on training Gray-MedNet using 3M publicly available gray-scale medical images including MRI, CT, X-ray, ultrasound, and PET. Furthermore, training Color-MedNet using 3M publicly available color medical images including histopathology, ophthalmology, and many other (See Fig.1). These images were collected from different repositories including Kaggle and IEEE, among others. Table 1 lists those repositories that will be used to train MedNet. These image repositories include 1) CT images (abdomen, bladder, brain, chest, kidney, cervix, breast), 2) MRI (neuroimaging, cardiovascular, liver, functional, oncology, phase contrast), 3) PET ( cardiology, infected tissues, neuroimaging, oncology, musculoskeletal, pharmacokinetics), 4) Histology ( epithelium, endothelium, mesenchyme, blood cells, neurons, germ cells, placenta), 5) X-Rays
( radiography, mammography, fluoroscopy, contrast radiography, arthrography, discography, dexa Scan)
6) Ultrasound(breast, doppler, abdominal, transabdominal, cranial, spleen, transrectal), 7) ophthalmic fundus images, 8) corneal topographic maps, 10) skin cancer images, among others. Data augmentation techniques and generative adversarial networks will be used to address the imbalance issue. It is worth mentioning that the datasets are collected, filtered, and prepared for training.

Once MedNet is trained, the fine-tuning process achieves to train MedNet on five small target datasets to check the effectiveness of MedNet in terms of transfer learning. Finally, MedNet will be publicly available to the research community to be specifically fine-tuned for dealing with any application of medical imaging.

## 5 MEDNET: CASE STUDY

A single experiment similar to MedNet has been implemented by our team on two medical image applications as explained in [1].

We employed two challenging medical imaging scenarios dealing with skin and breast cancer clas-sification tasks. Both tasks have a large dataset of images. To benefit from that, we used an archive of these tasks to improve the performance of recent datasets of the same tasks. Specifically, more than 200K unlabeled images of skin cancer were used to train the model. Then, it was fine-tuned by considering a small dataset of labeled skin cancers to classify them into two different classes, namely benign and malignant. Moreover, with the same training scenario, we have classified hematoxylin–eosin-stained breast biopsy images into four different classes: invasive carcinoma, in situ carcinoma, benign tumor, and normal tissue. In both tasks, we found that this technique of transfer learning from medical images improved the learning stage of the model and promote its generalization in a more suitable manner. Our TL technique improved the results better than the pre-trained models of ImageNet. Moreover, our team implemented different experiments which are similar to the MedNet idea with on single task of different medical applications [2,3,4,5,6]. Instead of building a pre-trained model for a single task, MedNet will be for most of the medical image tasks.

## 6 MEDNET: OUTCOMES

Although many applications from medical imaging achieved a high performance using the standard models (i.e. ImageNet), many of them failed when moving to clinical tests because still many drawbacks need to be addressed regarding using a small training set, generalization issues, and over-fitting. Now, MedNet can ease many of the negative effects derived from those pitfalls. In particular, MedNet provides the following outcomes:

Table 1. SOURCE DATASETS FROM DIFFERENT REPOSITORIES

| Title | link |
| --- | --- |
| Kaggle | https://www.kaggle.com/datasets |
| NIH | https://openi.nlm.nih.gov/ |
| Google Dataset | https://datasetsearch.research.google.com/ |
| 100 k Chest X-Rays | https://nihcc.app.box.com/v/ChestXray-NIHCC |
| The Cancer Imaging Archive (TCIA) | https://www.cancerimagingarchive.net/ |
| OASIS Brains Datasets | https://www.oasis-brains.org/ |
| Skin Lesion Analysis | https://www.isic-archive.com/ |
| Challenges of medical image analysis | https://grand-challenge.org/challenges/ |
| MedPix | https://medpix.nlm.nih.gov/home |
| FITBIR | https://fitbir.nih.gov/ |
| STARE | http://cecas.clemson.edu/ ahoover/stare/ |
| Cancer Digital Slide Archive | https://cancer.digitalslidearchive.org/ |
| UCI Machine Learning | http://archive.ics.uci.edu/ml/index.php |
| Computer Vision Online Image Archive | https://homepages.inf.ed.ac.uk/ |
| Lung Database | http://www.via.cornell.edu/crpf.html |
| RSNA AI Challenges | https://www.rsna.org/education/ai-resources-and-training/ai-image-challenge |
| Synapse | https://www.synapse.org/ |
| Retinal images | http://www2.it.lut.fi/project/imageret/diaretdb1/ |
| SICAS Medical Image | https://www.smir.ch/ |
| Various medical imaging datasets | https://github.com/sfikas/medical-imaging-datasets |

- MedNet will generalize very well.
- It is robust against over-fitting.
- It deals with the lack of training data from medical imaging.
- It guarantees that the model will learn the relevant features from medical images and minimize the efforts needed by the annotation process.
- It is not limited to a specific task, then it can be used in most of the applications of medical imaging. The learned features of MedNet can be used in medical image classification, segmentation, detection, disease prediction, and diagnosis, etc.
- It will be publicly available to the medical imaging research community both in Matlab and Python programming languages.

Paper summary: Training the CNN model on medical images instead of natural images (ImageNet) to learn relevant medical features. A huge number of medical images for training then fine-tune for the target small, labelled task such as cancer type classification.

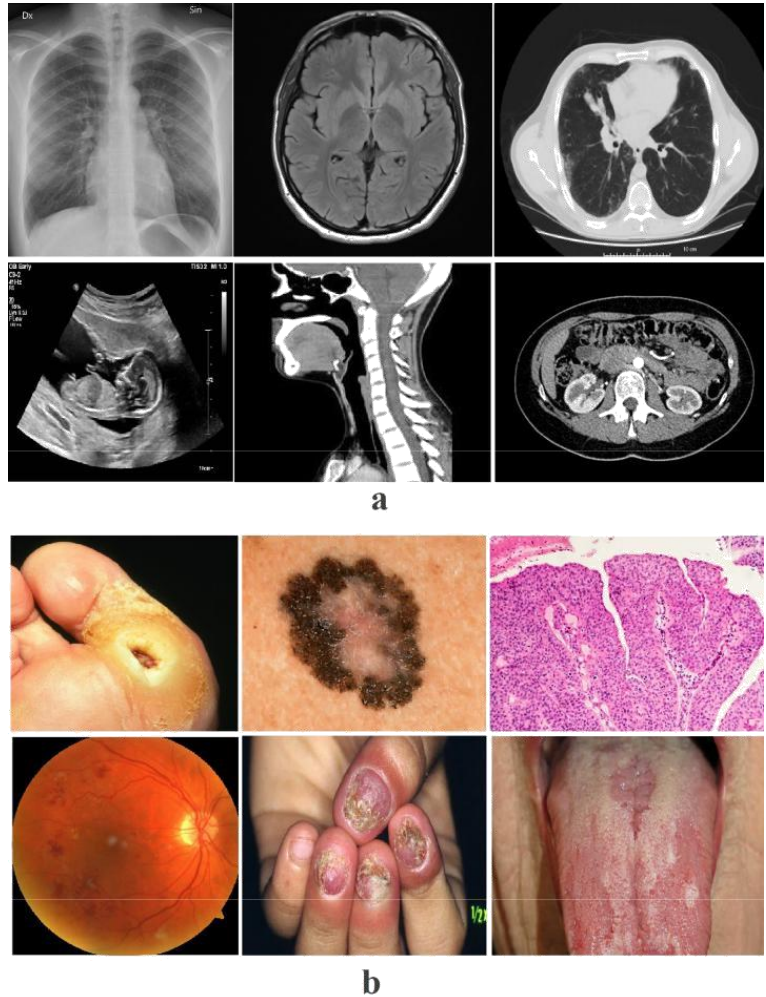

**Figure 1.** Medical Images, a) Medical images types to train Gray-MedNet, b) Medical image types to train Color-MedNet